\pdfoutput=1
\pdfoutput=1

\documentclass[11pt]{article}

\usepackage[final]{acl}

\usepackage{times}
\usepackage{latexsym}
\usepackage{amssymb}
\usepackage{amsmath}
\usepackage{todonotes}
\usepackage{floatrow}
\usepackage{multirow}
\usepackage{comment}

\newfloatcommand{capbtabbox}{table}[][\FBwidth]
\newtheorem{theorem}{Theorem}

\usepackage[T1]{fontenc}

\usepackage[utf8]{inputenc}

\usepackage{microtype}

\usepackage{inconsolata}

\usepackage{graphicx}

\title{Chain and Causal Attention for Efficient Entity Tracking}

\author{Erwan Fagnou \and Paul Caillon \and Blaise Delattre \and Alexandre Allauzen \\
        Miles Team, LAMSADE \\ Université Paris Dauphine-PSL \\ Paris, France
        \\
 \small{
   \textbf{Correspondence:} \href{mailto:erwan.fagnou@dauphine.psl.eu}{erwan.fagnou@dauphine.psl.eu}
 }
}

\begin{document}
\maketitle
\begin{abstract}
This paper investigates the limitations of transformers for entity-tracking tasks in large language models. We identify a theoretical constraint, showing that transformers require at least $\log_2 (n+1)$ layers to handle entity tracking with $n$ state changes. To address this issue, we propose an efficient and frugal enhancement to the standard attention mechanism, enabling it to manage long-term dependencies more efficiently. 
By considering attention as an adjacency matrix, our model can track entity states with a single layer.
Empirical results demonstrate significant improvements in entity tracking datasets while keeping competitive performance on standard natural language modeling. Our modified attention allows us to achieve the same performance with drastically fewer layers. 
Additionally, our enhanced mechanism reveals structured internal representations of attention. Extensive experiments on both toy and complex datasets validate our approach. Our contributions include theoretical insights, an improved attention mechanism, and empirical validation.
\end{abstract}

\section{Introduction}

Transformers \cite{vaswani2017attention} have deeply renewed state-of-the-art approaches in many
natural language processing (NLP) tasks. More specifically, the
attention mechanism can readily handle long-range dependencies and
complex linguistic phenomena. This kind of architecture 
surpasses the previous ones, like Markov models and recurrent
networks. For instance, transformers exhibit an expressivity in fluent
language generation, even for long text, that was not possible before.
Moreover, self-attention is very powerful in deriving rich and
meaningful representations of languages.

This major paradigm shift paved the way for Large Language Models (LLM). Starting with GPT \cite{radford2018improving} and BERT \cite{devlin2018bert},  many successors have emerged, witnessing a massive scaling up of model sizes by several orders of magnitude along with unprecedented improvements in performance. Nowadays, state-of-the-art approaches mostly rely on LLMs for most of the NLP, setting a new standard for the development and application of language models in both academia and industry \cite{brown2020language, liu2019roberta, raffel2020exploring, yang2019xlnet, he2021deberta}.

However, the amount of resources needed to train and use them has also increased by several orders of magnitude. This raises important concerns regarding their ecological impact and limits their use and applications \cite{luccioni2023estimating, faiz2024llmcarbon}. This sustainability issue has motivated different lines of work to make LLMs more frugal by optimizing architectures and reducing computational overhead without a drop in performance. Various approaches have been proposed, such as pruning, quantization, and efficient training techniques, to reduce the carbon footprint of LLMs \cite{dao2022flashattention, tay2022efficient, fedus2022switch, hinton2015distilling, hu2022lora, tay2022efficient}.

Most of these methods address this call for frugality in a task-agnostic way, while the complexity and computational resources may greatly differ depending on the downstream task under study. Some tasks like entity tracking imply long-range dependencies with chain-like memorization \cite{kim2023entity}.  
More precisely, for entity tracking, the model has to maintain and update the states of entities as described through a sequence of operations. While it looks fairly easy for humans, this is challenging for standard transformer models.
\citet{zhang-etal-2023-causal} indeed show that current LLMs have underwhelming performance on causal reasoning of event plausibility and entity states, despite the crucial part it plays in natural language reasoning.
Moreover, the complexity drastically increases with the sequence length. Therefore, in this work, we focus on the entity tracking task, which is critical in many applications like question-answering, dialogue systems, text-based games, and reasoning \citep{tandon-etal-2019-wiqa, mysore-etal-2019-materials, zhang-etal-2024-openpi2}.

In this paper, we introduce a modified attention layer to overcome this limitation. Our goal is to enhance the standard attention mechanism with a new design, to better cope with long-term and chain-like dependencies. For this purpose, our approach relies on a different interpretation of the attention matrix as an adjacency matrix.

Our contributions are the following:
\begin{itemize}
    \item We formalize entity tracking to prove that a transformer needs at least
    $\log_2 (n+1)$ layers to handle an entity tracking task with $n$ jumps, since the computation graph takes the form of a binary tree. This theorem is confirmed empirically.
    \item We design an improved attention mechanism to overcome this limitation with a single layer and show it works empirically for entity tracking tasks of arbitrary length.
    \item Our attention layer matches the performance of standard transformers on standard pre-training tasks (language, code), but greatly improves on specific entity tracking datasets.
\end{itemize}
Our theoretical and empirical results extend the understanding of how attention layers are used to handle entity tracking in language models. 

\section{Background and related work}

In this work, we focus on the decoder-only architecture~\cite{radford2018gpt} based on transformers~\cite{vaswani2017attention}. This kind of generative architecture is at the core of  Causal Language Modeling (CLM).  While there are many variants \cite{radford2018gpt, radford2019gpt2, brown2020gpt, raffel2020t5, lewis2020bart}, they all rely on the same framework. After a tokenization step of the texts, the embedding layer converts the pre-processed input in a sequence of vectors, \textit{i.e} a matrix. Then,  $L$ layers of transformers process this matrix, each made of a self-attention layer followed by a feedforward network. The output vectors are finally projected into logits for prediction purposes.

\subsection{Self-attention layer}
At the core of the transformer architecture is the attention layer. Given an input  $\mathbf{X} \in \mathbb{R}^{L\times d}$  ($L$ vectors of dimension $d$), a standard attention head outputs $\mathbf{Y}$, computed as follows:
\begin{align}
    \mathbf{A} &= \mathrm{Softmax}\left( \frac{(\mathbf{X} \mathbf{W_q}) (\mathbf{X} \mathbf{W_k})^T}{\sqrt{d_k}} + \mathbf{M}\right) \label{eq:attention} \\
    \mathbf{Y} &= \mathbf{A} \mathbf{X} \mathbf{W_v} = \mathbf{A}\mathbf{V} \label{eq:attention2}
\end{align}
where $\mathbf{W_q} \in \mathbb{R}^{d \times d_k}$, $\mathbf{W_k} \in \mathbb{R}^{d \times d_k}$, and $\mathbf{W_v} \in \mathbb{R}^{d \times d_v}$ are learned matrices. 
$\mathbf{M}$ is a masking matrix that enforces causality: a token cannot see a future token.  It is a strict upper-triangular matrix where the nonzero entries are $-\infty$.
Multi-head attention is obtained by stacking attention heads together, each with $d_k = d_v = d / n_{\text{heads}}$, and then performing a final multiplication by a learned matrix $\mathbf{W_o}$.

\paragraph{Multi-hop attention}
A work closely related to our findings is the Multi-Hop Attention Graph Neural Network (MAGNA) \cite{wang2021multihop}. Unlike standard graph neural networks (GNNs) that compute attention only between directly connected nodes, MAGNA incorporates multi-hop context information into attention computations at each layer, thereby increasing their receptive field. This enhancement of graph attention is similar to how we enhance attention for language modeling, yet shows significant differences in motivation, interpretation, and realization. 

\subsection{Entity tracking}
\label{ssec:entitytracking}
\begin{figure*}
\floatsetup[figure]{style=plain,subcapbesideposition=top}
\ffigbox{
\begin{floatrow}

\capbtabbox{%
    \begin{tabular}{|l|}
    \hline
    \textbf{Textual description} \\
    \hline
    The apple is in box A. \\
    There is nothing in box B. \\
    Move the contents of box A to box B. \\
    Put the book into box A. \\
    Move the contents of box B to box A. \\
    \hline
    What is in box A? \\
    \hline
    \end{tabular}
}{%
    \caption*{}
}

\capbtabbox{%
    \begin{tabular}{|l|}
    \hline
    \textbf{Abstract} \\
    \hline
    $x_0 \leftarrow \{\text{apple}\}$ \\
    $x_1 \leftarrow \{\text{}\}$ \\
    $x_2 \leftarrow x_0 \cup x_1 ,\;\; x_3 \leftarrow \{\}$ \\
    $x_4 \leftarrow x_3 \cup \{\text{book}\}$ \\
    $x_5 \leftarrow x_2 \cup x_4$ \\
    \hline
    $x_5 = \;?$ \\
    \hline
    \end{tabular}
}{%
   \caption*{} 
}

\hspace{0.4cm}
\ffigbox[\FBwidth]{
    \includegraphics[width=3.3cm]{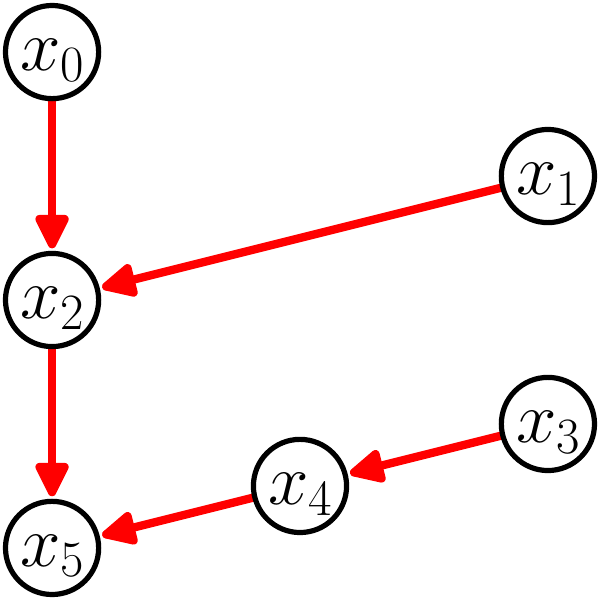}
}{%
   \caption*{} 
}
\end{floatrow}
}{
\vspace*{-0.8cm}
\caption{Illustration of an entity tracking task and its computational graph representation. From left to right: (a) Textual description of the task. (b) Abstract representation of the task in terms of variable assignments. (c) Graph representation of the task showing dependencies between states.}
    \label{fig:entity_tracking_task}
}
\end{figure*}

Entity tracking, also referred to as state tracking, is the task of keeping track of the state of one or several entities over time. Such a problem may happen in a variety of ways. For example, consider a cooking recipe: "Place the sliced onions in a frying pan. Stir until golden brown, then transfer them to a plate.". Here, the model needs to comprehend that the onions are now cooked and on the plate.
Figure~\ref{fig:entity_tracking_task} provides another illustration of this task. 

\paragraph{Entity tracking and interpretability}
The study of entity tracking in language models started with the work of \citet{li2021implicit}. They transformed two existing datasets, Alchemy \cite{long2016simpler} and TextWorld \cite{cote2019textworld}, into full texts. Inputs from the Alchemy dataset start by describing the contents of some beakers (e.g. "The first beaker has 1 green, the second beaker has 2 red, the third beaker has 3 red."), and then describe changes made to these beakers ("Pour the last red beaker into beaker 1. Mix."). We then expect the model to keep track of what is inside each beaker. With two LLMs, T5~\cite{raffel2020t5} and BART~\cite{lewis2020bart}, they investigated what information is stored in the hidden states and where. They conclude that the internal representations of entity states are highly interpretable and structured.

Similarly, \citet{li2023othello}  investigated the contents of the hidden states in a GPT model \cite{radford2018gpt, brown2020gpt} trained on Othello game records. They showed that the board state can be recovered from the hidden states. It shows that the LLM maintains a high-level representation of the game board. Furthermore, \citet{toshniwal2022chess} showed the same fact with  GPT-2 models \cite{radford2019gpt2} on chess games. Therefore, transformer models can achieve impressive performance when trained on entity tracking tasks, but this ability has a cost (see Theorem~\ref{theorem_limitation}). 

\paragraph{Emergence of entity tracking abilities} Identifying the conditions that enable language models to acquire entity-tracking capabilities is a recent line of work.
In \cite{kim2023entity}, pre-trained LLMs like  GPT-3, GPT-3.5, and Flan-T5 \cite{chung2022flant5} are evaluated on entity tracking tasks without fine-tuning. Their empirical results suggest that pre-training on text corpora only is not enough to build skills for entity tracking. Notably, only models additionally trained on code data demonstrated reliable entity tracking performance. Further emphasizing the role of training data, \citet{kim2024code} investigated various training datasets and concluded that entity tracking abilities predominantly emerge when models are trained on code-related data. 

In a related study, \citet{muennighoff2023scaling} explored the impact of integrating code-based pre-training into language models, and observed that incorporating code data up to 50\% during pre-training significantly enhances long-term state tracking capabilities without compromising 
performance on traditional natural language modeling tasks.
In  \cite{prakash2024fine-tuning}, the authors explore how fine-tuning can modify transformer models' mechanisms for entity tracking. They observe a clear improvement without fundamentally altering the mechanistic operation of the model. 

All of these empirical results demonstrate the potential benefits of incorporating domain-specific knowledge into transformer models to improve their adaptability and performance in tasks that require persistent state tracking. However, they focus on the importance of providing adapted training data, while our goal is to modify the attention mechanism to leverage this ability.

\section{Theoretical limits of transformers for entity tracking}

Following the introduction of entity tracking in section \ref{ssec:entitytracking}, we here more formally define this task to specify the theoretical limitations of transformers. 

\subsection{Entity tracking as a graph traversal task}

An instance of entity tracking starts by initializing the states of some entities, followed by a sequence of operations that modify these states. Each operation can be written as a function $f_i$, applied to the states of the entities $(x_0, \dots, x_{i-1})$, and which produces the new state $x_i$ of one of the entities:
\begin{equation}
    x_i \leftarrow f_i(x_0, \dots, x_{i-1})
\end{equation}

An example is provided in Figure~\ref{fig:entity_tracking_task}, where $x_0$ and $x_1$ are the initial states of boxes A and B respectively, and $x_2$ represents the new state of box B after receiving the contents of box A.

\paragraph{Representation as a computational graph} \label{sec:graph_definition}
Now that we can essentially represent an entity tracking problem as an algorithm, we can consider the dependency graph between the variables. The nodes are the variables $x_i$, and there is a directed edge from $x_i$ to $x_j$ if $x_i$ has been used directly to compute $x_j$. As a variable can only be computed from previous variables, the graph is a Directed Acyclic Graph. We show how we convert an entity tracking task example into its equivalent computational graph in Figure~\ref{fig:entity_tracking_task}.

\subsection{A theoretical limitation of transformers}

We now show that transformers have inherent limitations when dealing with tasks that require deep and complex state tracking, formalized in the following theorem:

\begin{theorem} \label{theorem_limitation}
    Given an entity tracking task instance, let $\mathcal{G}$ be its corresponding computational graph as defined in Section~\ref{sec:graph_definition}.     
    We assume that each attention layer has a receptive field equal to 1 in the computational graph \footnote{In other words, a layer cannot make the connection between two nodes of the computational graph that are not directly connected. We further discuss the motivation and validity of this assumption in Appendix~\ref{sec:assumption_discussion}.}.
    
    Then a transformer model requires a minimum of $L_{\min} (\mathcal{G})$ attention layers to solve the task, with:
    \begin{equation}
        L_{\min} (\mathcal{G}) = \lceil \log_2 ( \text{depth}(\mathcal{G}) + 1 ) \rceil
    \end{equation}
    where we define $\text{depth}(\mathcal{G})$ as the length of its longest path. 
\end{theorem}

This theorem implies that the number of layers required grows logarithmically with the number of state transitions, due to the inherent structure of the dependency graph in such tasks. This is particularly relevant for long sequences where many state changes occur.

\begin{figure}[t]
  \includegraphics[width=\columnwidth]{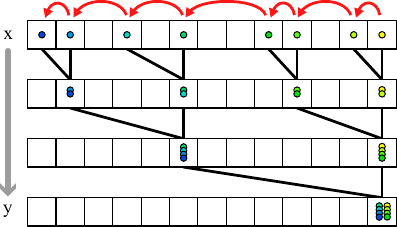}
  \caption{Illustration of how a standard transformer can process a sequence containing chained dependencies. Red arrows represent the reference to a previous node, and black lines show attention connections. With $8$ nodes (represented as colored circles), $log_2(8) = 3$ attention layers are needed to process and gather all the information.}
  \label{fig:binary_graph}
\end{figure}

This result comes naturally by considering a single chain of dependencies between variables. We illustrate this intuition in Figure~\ref{fig:binary_graph}. Similarly to a linked list, an element initially only knows about the previous element. The first attention layer then allows it to look one step further. As we apply more attention layers, a binary tree structure emerges, as shown in black lines in Figure~\ref{fig:binary_graph}. We present a more detailed proof in Appendix~\ref{sec:theorem_proof}. 

\section{Method}

In this section, we introduce a novel attention mechanism that enhances traditional attention by enabling a single attention layer to handle long-term entity tracking.

\subsection{Enhanced attention}

With the notations of Equations \ref{eq:attention} and \ref{eq:attention2}, a standard attention layer outputs \(\mathbf{Y}=\mathbf{A}\mathbf{V}\). While effective, this approach can struggle with capturing long-range dependencies due to the limited receptive field of a single layer, which is linked to the way the computational graph is explored. To address this, we propose an enhanced attention mechanism that uses a modified attention map to extend the receptive field, allowing the model to capture long-term dependencies more effectively.

\paragraph{Adjacency matrix perspective}
The attention matrix from the first layer can be viewed as an adjacency matrix representing dependencies in a computational graph. By leveraging the properties of the adjacency matrix, our approach simplifies entity tracking to a single layer instead of requiring multiple layers.
Consider the adjacency matrix \(\mathbf{A}\) of a graph, where each entry represents the direct connection between nodes (i.e., the attention scores). The power of this matrix, \(\mathbf{A}^2\), captures paths of length 2, \(\mathbf{A}^3\) captures paths of length 3, and so on. To fully understand the dependencies and track states in a computational graph where the longest chain is \(n\), it is necessary to consider paths of all lengths up to \(n\).
This can be achieved by the matrix series: \(\mathbf{A} + \mathbf{A}^2 + \mathbf{A}^3 + \cdots\)
which represents the sum of all possible paths. Assuming this series converges, it has a closed form formula: \(\mathbf{A} (\mathbf{I} - \mathbf{A})^{-1}\). Integrating this into the attention layers allows the model to capture paths of any length within the computational graph, extending its ability to track states over long dependencies.

\paragraph{Chain and Causal Attention Layer (ChaCAL)}
Building upon this intuition, we propose the enhanced attention mechanism ChaCAL where the output is computed as:
\begin{equation} \label{eq:layer_out}
    \mathbf{Y} = (1 - \gamma) \cdot \mathbf{A}(I - \gamma \mathbf{A})^{-1}\mathbf{V}
\end{equation}
We introduce a new parameter \(\gamma \in [0,1)\) which has two roles: \textit{i)} ensuring convergence of the series $\mathbf{A} + \gamma\mathbf{A}^2 + \gamma^2\mathbf{A}^3 + \cdots$, and \textit{ii)} allowing continuous interpolation between standard attention ($\gamma=0$, reverting the expression to Equation~\ref{eq:attention2}) and ChaCAL ($\gamma \approx 1$). We analyze the influence of $\gamma$ on performance and stability in Appendix~\ref{sec:gamma_study}, concluding that any value between 0.8 and 1 seems to work, and that 0.9 is a good default choice. Additionally, in practice we remove the diagonal part of the adjacency matrix in the inverse, which corresponds to a token attending to itself, as it is trivial and improves learning.
This enhanced mechanism enables the model to consider longer paths, thereby improving its ability to handle long-term dependencies.

\subsection{Fixed point interpretation}

Previous works have demonstrated success in replacing multiple transformer layers with a single layer repeated multiple times \cite{reid2021subformer, wang2024residual} or even finding its fixed point \cite{bai2019deq}. Our algorithm can be interpreted as finding the fixed point of the following linear function:
\begin{equation}
    f(\mathbf{Z}) = \mathbf{A} \cdot (\gamma \mathbf{Z} + (1 - \gamma) \mathbf{V})
\end{equation}
While the absence of nonlinearity in $f$ limits the expressiveness of this layer, the benefit lies in the computationally efficient closed-form solution provided by Equation~\ref{eq:layer_out}. This approach enables the model to handle long-term dependencies with a single attention layer.

\subsection{Efficient computation}

Considering the context of causal language modeling, the attention matrix $\mathbf{A}$ is lower-triangular, and so is $\mathbf{I}-\gamma \mathbf{A}$. Moreover, as long as $\gamma < 1$, its eigenvalues are nonzero, meaning it is invertible. We can thus write the output $\mathbf{Y}$ as the solution of:
\begin{align} \label{eq:tri_system}
    &\mathbf{B}\mathbf{Y} = \mathbf{C} \\
    \text{where}&\quad \Biggl\{ \begin{array}{l} \mathbf{B} = \mathbf{I} - \gamma \mathbf{A} \\ \mathbf{C} = (1 - \gamma) \cdot \mathbf{A}\mathbf{V}  \end{array} \nonumber
\end{align}
This has a unique solution and is easy to solve as it is a triangular system. Note that it does not require explicitly computing the inverse of the matrix, which leads to a more stable and efficient implementation. The algorithm however differs between training and inference, as inference involves decoding one token at a time.

\paragraph{Training} The solution to Equation~\ref{eq:tri_system} can be computed efficiently using any triangular solver. We use \texttt{torch.linalg.solve\_triangular} from the PyTorch library, which runs efficiently on GPU and automatically handles differentiation.

\paragraph{Inference} When decoding, we solve one row of the triangular system per token decoded. This is done by forward substitution:
\begin{equation}
    y_t = \frac{1}{\mathbf{B}_{tt}} \left[\mathbf{C}_t - \sum_{i=1}^{t-1}\mathbf{B}_{m,i}\;y_i \right]    
\end{equation}

where $y_t$ denotes the $t$-th row of $Y$, i.e. the output vector at position $t$ in the sequence.

More generally, if we have already solved the system for the first $t$ tokens, and we want to solve the next $k$ tokens, we can use:
\begin{align} \label{eq:tri_system_decomp}
    &\mathbf{B}\mathbf{Y} = \mathbf{C} \nonumber\\
    \Leftrightarrow& \begin{bmatrix} \mathbf{B}_0 & 0 \\ \mathbf{B}_1 & \mathbf{B}_2 \\ \end{bmatrix} \begin{bmatrix} \mathbf{Y}_{1:t} \\ \mathbf{Y}_{t:t+k} \\ \end{bmatrix} = \begin{bmatrix} \mathbf{C}_{1:t} \\ \mathbf{C}_{t:t+k} \\ \end{bmatrix} \nonumber\\
    \Leftrightarrow& \begin{cases} \mathbf{B}_0 \mathbf{Y}_{1:t} = \mathbf{C}_{1:t} \\ \mathbf{B}_2 \mathbf{Y}_{t:t+k} = \mathbf{C}_{t:t+k} - \mathbf{B}_1 \mathbf{Y}_{1:t}\end{cases}
\end{align}

In other words, Equation~\ref{eq:tri_system_decomp} means that we can encode the first $t$ tokens (e.g. a text prompt) by solving a first triangular system, and then use the solution to compute the next tokens efficiently with a second triangular system.

\section{Experiments}

Throughout all of our experiments, we base our models on the GPT-2 \cite{radford2019gpt2} architecture, which is the base of most modern LLMs. We compare standard transformer models and our proposed enhanced attention layer on 3 experiments: an abstract toy dataset, a textual entity tracking task, and fine-tuning on a standard language modeling task. 
While our method introduces a new hyperparameter $\gamma$, we used $\gamma=0.9$, as we show in Appendix~\ref{sec:gamma_study} that any value between 0.8 and 1 seems to work fine. Detailed experimental information is in Appendix \ref{sec:ExpSetup}.

\subsection{Experiment 1: Toy dataset}
\label{sec:toy_dataset}

We first confirm the assumptions from previous sections by experimenting on a toy dataset, designed to highlight the limitation of transformers for long-term entity tracking.

\subsubsection{Setup}

\paragraph{Task description}
We generate sequences of numbers $[x_1, \dots, x_n]$. The first $k$ tokens contain random values. The next $k$ tokens contain the shuffled indices of the first $k$ tokens. We then repeat the process: the next $k$ tokens contain the shuffled indices of the previous $k$ tokens, and so on.

The goal of the model is to predict, for every position k, the indices included in the reference chain: $[x_k, x_{x_k}, x_{x_{x_k}}, \dots]$.

\paragraph{Goals}
We define this abstract toy task to experimentally show the limitations of attention in a controlled and arbitrarily difficult setting.
In our experiments, we take sequences of length $n = 128$ ($16$ blocks of size $k=8$). Therefore, the longest reference chain will be of length $15$, and we will expect from Theorem~\ref{theorem_limitation} that a transformer needs at least $L=\log_2(16)=4$ layers to solve the task.

\subsubsection{Results}

\begin{figure}[t]
  \includegraphics[width=\columnwidth]{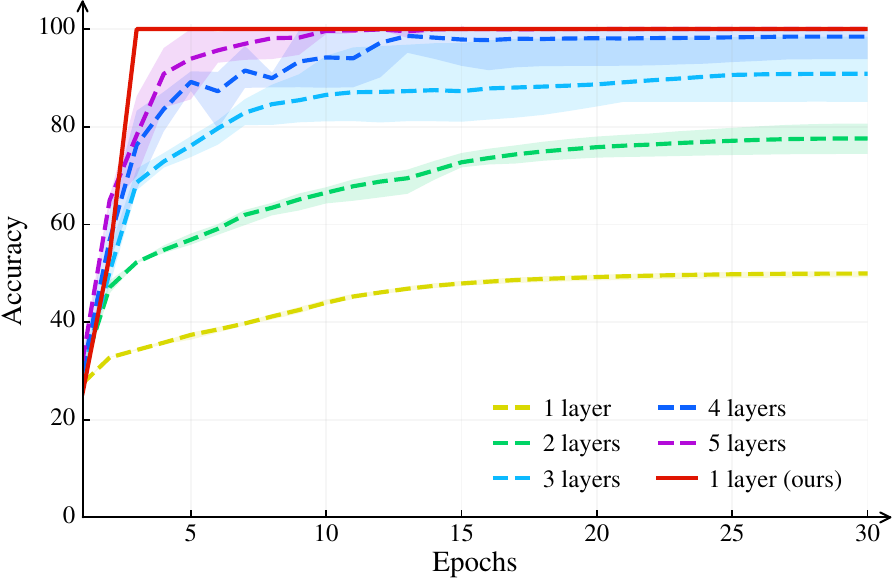}
  \caption{Test accuracy of transformer models during training on our toy dataset. The standard transformer architecture struggles to learn the task and needs 4 to 5 layers to reach 100\%, while our enhanced attention consistently solves the task with only one layer. We show the average, min and max values over 4 runs for each model.}
  \label{fig:toy_task_curves}
\end{figure}

\begin{table}
    \centering
    \begin{tabular}{l|crcc}
    \hline
    \textbf{Attention}                 & \textbf{\boldmath{$L$}} & \textbf{Size} & \textbf{Loss} & \textbf{Accuracy}  \\
    \hline
\multirow{5}{*}{Standard}
             & 1 & 3.3M & 1.20 & 49.9\% $\pm$ 0.5 \\
             & 2 & 6.4M & 0.54 & 77.6\% $\pm$ 2.8 \\
             & 3 & 9.6M & 0.21 & 90.9\% $\pm$ 5.2 \\
             & 4 & 12.7M & 0.03 & 98.5\% $\pm$ 2.7 \\
             & 5 & 15.9M & 0.00 & 100\% $\pm$ 0.0 \\
 \hline
ChaCAL       & 1 & 3.3M & \textbf{0.00} & \textbf{100\%} $\pm$ 0.0 \\
    \hline
    \end{tabular}
    \caption{Test loss and accuracy for the models evaluated on the toy dataset. We compare standard transformers with $L=1$ to 5 layers, to a single transformer layer using our enhanced attention.}
    \label{tab:toy_task_table}
\end{table}

Results for the toy dataset are shown in Figure~\ref{fig:toy_task_curves} and Table~\ref{tab:toy_task_table}. We evaluate standard transformer models following the GPT-2 architecture \cite{radford2019gpt2}, with 1 to 5 layers, as well as our proposed enhanced attention with a single layer. Note that all runs use the same setup and hyperparameters, and only the architecture is modified. 

A first observation is that standard transformers need at least $L=4$ layers to get a close to perfect score, which is the theoretical minimum number of layers predicted by Theorem~\ref{theorem_limitation}.  However, $L=5$ layers are needed to solve the task consistently.

ChaCAL achieves a perfect score of 
  100\%  on test accuracy in only 3 epochs, and with a single layer.
 This synthetic dataset, specifically designed for difficult entity-tracking tasks, showcases that our architecture is particularly effective in handling such challenges.

\subsection{Experiment 2: Boxes dataset}

We base our second experiment on the dataset designed by \citet{kim2023entity} and slightly modified by \citet{prakash2024fine-tuning}. The task involves placing items in boxes and moving them around, to determine where the items are after all these operations correctly.

\subsubsection{Setup and Task Description}

Input from this dataset starts by describing the contents of the boxes (e.g. "The camera is in Box F, the gift is in Box D."). Then a sequence of operations follows. There are 3 possible operations:
\begin{itemize}
    \item \textbf{"Put"}: some items are placed in a box (e.g. "Put the shirt into Box H.")
    \item \textbf{"Remove"}: some items are removed from a box (e.g. "Remove the camera from Box F.")
    \item \textbf{"Move"}: some items are moved from one box to another. It can be formulated either explicitly (e.g. "Move the shirt from Box H to Box F."), or implicitly (e.g. "Move the contents of Box H to Box F.") which requires the model to know what was in the box before.
\end{itemize}

Two versions of this dataset are used in our experiments. The \textit{default} version is the original dataset designed by \citet{kim2023entity} with the minor modifications from \citet{prakash2024fine-tuning}, except we increase the number of operations per example from 12 to 32, and we train the model to predict the contents of all boxes instead of only one.

The \textit{advanced} version always uses the implicit formulation of the "Move" operation, and enforces that only half boxes contain items, which means each item will undergo more operations. These modifications make the computational graph of the entity tracking much deeper, which is harder to handle with standard transformers according to Theorem~\ref{theorem_limitation}). Additionally, we randomize the number of operations per example using a log-uniform distribution between 1 and 31, in order to include more easy examples. We observed this was important for making all the models learn the task faster and more consistently.

\paragraph{Models}
Solving the task requires a two-step process: first understanding what each sentence means, and then reasoning about these sentences to deduce the contents of each box. This cannot be done in a single layer, so we added a standard transformer layer before our ChaCAL. 

\subsubsection{Results}

\begin{figure}[t]
  \includegraphics[width=\columnwidth]{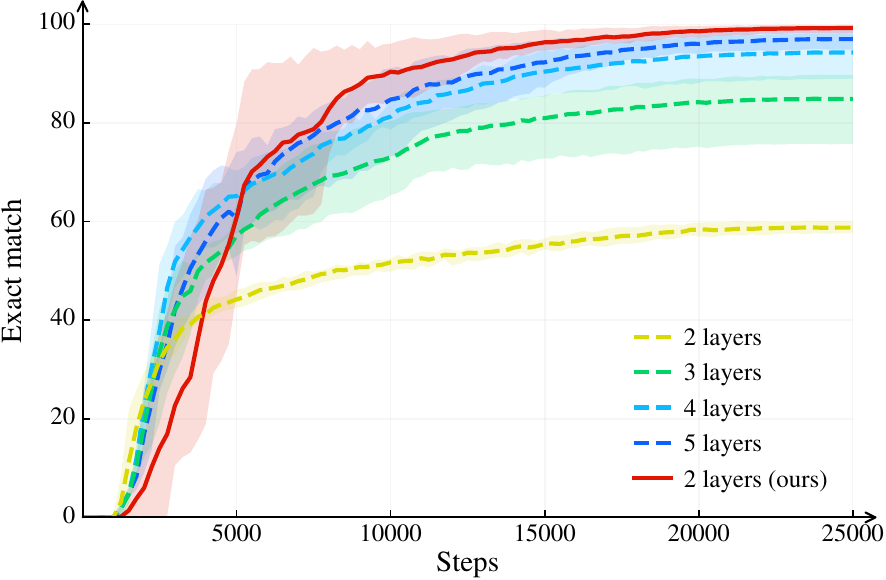}
  \caption{Exact match rate on the test set of each model during training on the \textit{advanced} version of the boxes dataset. }
  \label{fig:boxes_curves}
\end{figure}

\begin{table*}
    \centering
    \begin{tabular}{l|crcclccc}
    \hline
    
  & & & \textbf{Training} & \multicolumn{2}{c}{\textbf{Default}} & & \multicolumn{2}{c}{\textbf{Advanced}} \\ \cline{5-6} \cline{8-9}
    \textbf{Attention layer}                 & \textbf{\boldmath{$L$}} & \textbf{Size} & \textbf{time} & \textbf{Loss} & \textbf{Exact match} & & \textbf{Loss} & \textbf{Exact match} \\
    \hline
\multirow{4}{*}{Standard}
             & 2 & 6.4M  & $\times 1.00$  & 0.0543 & 45.1\% $\pm$ 2.5 && 0.0626 & 58.8\% $\pm$ 1.2 \\
             & 3 & 9.6M  & $\times 1.46$  & 0.0316 & 65.3\% $\pm$ 19.2 && 0.0198 & 84.8\% $\pm$ 6.4 \\
             & 4 & 12.7M & $\times 1.90$  & 0.0129 & 83.7\% $\pm$ 1.6 && 0.0075 & 94.3\% $\pm$ 4.0 \\
             & 5 & 15.9M & $\times 2.37$  & \textbf{0.0102}  & \textbf{84.6\% $\pm$ 5.2}  && 0.0040 & 97.0\% $\pm$ 1.7 \\
 \hline
ChaCAL (ours)  & 2 & 6.4M & $\times 1.36$  & 0.0234 & 70.6\% $\pm$ 13.8 && \textbf{0.0012} & \textbf{99.1\% $\pm$ 0.7 } \\
    \hline
    \end{tabular}
    \caption{Test loss and accuracy for the models evaluated on the \textit{default} and \textit{advanced} versions of the boxes dataset. We compare standard transformers with $L=2$ to 5 layers, to a 2-layered transformer using our enhanced attention in the second layer.}
    \label{tab:boxes_task_table}
\end{table*}

We report our results for both versions of the dataset in Table~\ref{tab:boxes_task_table}. In particular, we show the \textit{exact match} rate, i.e. the proportion of the test set for which the model generates the same sequence as the ground truth. In other words, this metric measures the ability of the model to track the content of all boxes perfectly.

On the \textit{default} version of the dataset, ChaCAL outperforms the standard transformer with $L=2$ or 3 layers, but does not improve over 4 or 5 transformer layers. Our intuition is that this dataset is too easy, as most of the sentences explicitly mention the items that are added, moved, or removed, which makes the contents of the boxes easy to track. Additionally, we observe that all models struggle to learn this task quickly, which may be caused by the sequences being too long, leading to slow convergence and high variance.

The \textit{advanced} version, however, highlights more the advantages of ChaCAL over standard transformer models. We show the learning curves of each model on this dataset version in Figure~\ref{fig:boxes_curves}. Not only does it solve the task, but it also does it much more efficiently in comparison to what would be needed with standard transformers. As shown in Table~\ref{tab:boxes_task_table}, ChaCAL is 1.75 times faster than a 5-layered transformer, but still beats it by 2\% (99.1\% against 97.0\%).

\subsection{Experiment 3: Language modeling}

In this final experiment, we evaluate ChaCAL on a standard natural language modeling task. Starting from the pre-trained GPT-2 model, we replace its attention with ours and fine-tune the model. As the pre-training dataset of GPT-2 was not released publicly, we use a subset of OpenWebText \cite{gokaslan2019openwebtext} -- its open-source version -- to fine-tune both the standard and modified models. 

\paragraph{Goal} Our experimental setup is highly unbalanced towards the standard transformer model, as we initialize the weights with those of a model pre-trained with standard attention. Nevertheless, our goal with this experiment is to assess the viability of our modified attention layer for standard natural language processing tasks.

\paragraph{Results}

\begin{table}
    \centering
    \begin{tabular}{l|crcc}
    \hline
    \textbf{Model}                 & \textbf{Perplexity} \\
    \hline
GPT-2 (pre-trained)        & 23.78  \\
GPT-2 (fine-tuned)         & 20.15  \\
 \hline
GPT-2 + ChaCAL (fine-tuned)  & 21.46  \\
    \hline
    \end{tabular}
    \caption{Perplexity of each model on a subset of the OpenWebText dataset, trained for causal language modeling.}
    \label{tab:lm_task_table}
\end{table}

As shown in Table~\ref{tab:lm_task_table}, a short fine-tuning of a pre-trained model with our new attention layer can recover a perplexity similar to the model fine-tuned with standard attention layers.

\section{Discussion}

Our results correlate with the theoretical prediction that transformers require at least $\log_2 (n+1)$ layers for entity tracking tasks with $n$ jumps. Indeed, our toy dataset has $n=15$ and models only start reaching 100\% accuracy at $L=4=\log_2(16)$. Similarly, our advanced version of the boxes dataset has $n \leq 31$ and we notice that $L=5=\log_2(32)$ layers are needed to get a perfect score. This confirms that \textit{i)} transformers are indeed limited for long-term entity tracking tasks, and \textit{ii)} they can successfully learn the task even when they barely have enough layers.

The ChaCAL model, with our improved attention mechanism, successfully learns entity tracking tasks using the minimal number of layers, outperforming by far standard transformers on these specific datasets. This demonstrates the frugality and efficiency of our method, as it manages complex tasks with fewer resources compared to standard transformers. We however notice a slower convergence at the beginning of training and a greater variance in the training curves. We believe that ChaCAL could greatly benefit from understanding better its training dynamics (e.g. a more suited initialization, or regularization), which we leave as future work.

While our modified attention mechanism does not significantly enhance general language modeling tasks, this observation is in line with several previous studies, such as \citet{kim2023entity}, \citet{kim2024code}, and \citet{muennighoff2023scaling}. Entity tracking is a relatively rare subtask within language modeling, indicating that pre-training on natural language alone does not foster robust entity-tracking capabilities. We also explain the absence of immediate benefits in language modeling by the large size of the models, which have an abundance of layers that allow them to handle most entity tracking tasks in theory. As ChaCAL was specially designed to reduce the number of attention layers, one may try to explore other architectural choices more suited to our purpose, involving shallower but wider models for instance.

A notable finding is that our model with $\gamma = 0.9$ across all layers reaches performances similar to standard transformers. Instead of fine-tuning a model that was pre-trained using standard attention, pre-training from scratch with our novel attention mechanism could lead to a more efficient attention structure optimized for entity tracking.

Finally, the high computational efficiency of our method mainly relies on the attention matrix being lower-triangular, which is only the case in decoder transformers for causal language modeling. Further research may attempt to generalize ChaCAL to non-causal transformers such as BERT \cite{devlin2018bert} and many others \cite{vaswani2017attention, liu2019roberta, he2021deberta, raffel2020t5}.

\section{Conclusion}

In this study, we have theorized that transformers require at least $\log_2 (n+1)$ layers for effective entity tracking tasks with $n$ jumps. Multiple experiments on entity tracking tasks, with both a toy dataset and a more complex natural language dataset, validate our claims and consistently demonstrate that transformers need at least this many layers to achieve perfect performance.

The ChaCAL model, leveraging our novel attention mechanism, excels in entity tracking tasks while drastically lowering layer requirements compared to standard transformers, and without adding more parameters nor requiring special tuning. This underscores the frugality and computational efficiency of our approach.
At last, fine-tuning experiments of GPT-2 with our enhanced attention on a natural language modeling task suggests that ChaCAL can at least match the standard transformer models.

Several key questions emerge: how does pre-training from scratch with ChaCAL compare to fine-tuning existing models? What are the optimal settings for $\gamma$ across different layers and tasks, and can it be dynamically adjusted during training to enhance the model's adaptability and performance?

In conclusion, this study advances our understanding of attention mechanisms in transformers, particularly in optimizing models for entity-tracking tasks. By bridging theory with practical insights, we lay a foundation for future research aimed at developing more efficient transformer architectures tailored to specific application domains.

\section*{Limitations}

Firstly, our experiments on Language Modeling tasks were limited in scope. While our attention mechanism showed promising results for entity tracking, its impact on general language modeling tasks, beyond those specifically designed to evaluate entity tracking, remains understudied. Future work should explore a broader range of language tasks to assess the generalizability and versatility of our approach.

Secondly, while the theoretical foundation of our layer requirement prediction is agnostic to specific transformer architectures, we primarily evaluated our ChaCAL model on GPT-2 layers. Extending this evaluation to other transformer architectures, such as GPT-3, BERT, or custom-designed models, would provide a more comprehensive understanding of the applicability and performance of our attention mechanism across different frameworks.

Additionally, our study fixed $\gamma = 0.9$ uniformly across all layers. Exploring adaptive strategies for $\gamma$, including layer-specific or task-specific tuning during training, could potentially enhance the model's flexibility and performance across a more diverse set of tasks.

Moreover, our experiments focused primarily on entity tracking in natural language datasets. Extending the evaluation to other domains such as code understanding, mathematical reasoning, and summarization tasks would elucidate the broader applicability of our attention mechanism in various real-world applications.

Lastly, while we discussed the computational frugality of our ChaCAL model compared to standard transformers, further investigation into the computational efficiency and scalability of our approach on larger datasets and models is necessary to fully assess its practical implications.

Addressing these limitations in future research endeavors will contribute to refining and validating the effectiveness of our attention mechanism in enhancing transformer architectures for a wide range of tasks and domains.


\section*{Acknowledgments}
This work was performed using HPC resources from GENCI-IDRIS (grants AD011015154 and A0151014627), and received funding from the French National Research Agency (ANR SPEED-20-CE23-0025) and the French Government via the program France 2030 ANR-23-PEIA-0008, SHARP.

\bibliography{bibliography}

\appendix

\section{Proof of Theorem~\ref{theorem_limitation}}
\label{sec:theorem_proof}

Consider an entity tracking task represented by a computational graph \( \mathcal{G} \), where nodes represent state variables and edges denote dependencies between states. Let \( \text{depth}(\mathcal{G}) \) denote the length of the longest path in \( \mathcal{G} \).

We start by identifying the longest path \( P = (v_1, v_2, \ldots, v_n) \) in \( \mathcal{G} \) with \( n = \text{depth}(\mathcal{G}) \).

Each layer of the transformer model has a receptive field of 1 in \( \mathcal{G} \), allowing it to propagate information only between directly connected nodes. 

To track dependencies across the longest path \( P \), we proceed as follows:

\paragraph{Base Case:}

Initially, the transformer initializes hidden states based on the input, encoding local information at each node \( v_i \) along path \( P \).

\paragraph{Inductive Step:}

Assume after \( L \) layers, the model can integrate information across paths of length \( 2^L-1 \) along path \( P \).

At \( L+1 \) layers, each node \( v_i \) on path \( P \) can access information up to \( 2^{L+1} - 1 \) steps away. This is because:
\begin{itemize}
    \item \( v_i \) already contains information about \( (v_i, \ldots, v_{i+2^L-1}) \),
    \item \( v_{i+2^L} \) contains information about \( (v_{i+2^L}, \ldots, v_{i+2^{L+1}-1}) \).
\end{itemize}

Using the attention layer to query the contents of \( v_{i+2^L} \), information from both \( v_i \) and \( v_{i+2^L} \) can be aggregated, enabling the next hidden state associated with \(v_i\) to contain information about \( (v_i, \ldots, v_{i+2^{L+1}-1}) \). Because of our assumption, we know that no more information could have been aggregated.

Therefore, \( L_{\min}(\mathcal{G}) = \lceil \log_2 (n + 1) \rceil \) layers are necessary for the transformer model to effectively handle dependencies along the longest path \( P \) in \( \mathcal{G} \).

\section{Motivation and validity of the assumption in Theorem~\ref{theorem_limitation}}
\label{sec:assumption_discussion}

The key assumption we make in Theorem~\ref{theorem_limitation} is that each attention layer only has a receptive field of 1 in the dependency graph. The motivation is that if a node only contains information about the previous node, then it has no way of directly accessing nodes that are 2 or more jumps away. As such, the only node it can attend to in the attention layer is the previous node. This is analogous to a linked list in programming, where each cell only contains a pointer to the next cell, and does not know where the other cells are stored in memory.

Using attention, a node A can only query information from the previous node B. After one iteration, using the new information from node B about the next node C, it is then possible to query information from node C about node D, and so on. In a sequential algorithm, this would require $\mathcal{O}(L)$ steps, $L$ being the chain length. However, transformers process each token in parallel, which leads to $\mathcal{O}(\log L)$ steps instead, and which is what we prove more formally in Theorem~\ref{theorem_limitation}.

A limit of the assumption is that in case the model overfits, the model might manage to "skip" nodes and look at nodes further away. Indeed, it could use attention to gather information about the whole sequence, and then use the feedforward layer to sort out and process the useful information. This is never optimal unless the model is largely overparametrized. This complicates the formalization of the assumption. Still, in practice we did not observe such phenomenon in our entity-tracking experiments, and out results correlate very well with the predictions of the theorem.

\section{Influence of $\gamma$}
\label{sec:gamma_study}

\begin{figure}[h]
  \includegraphics[width=\columnwidth]{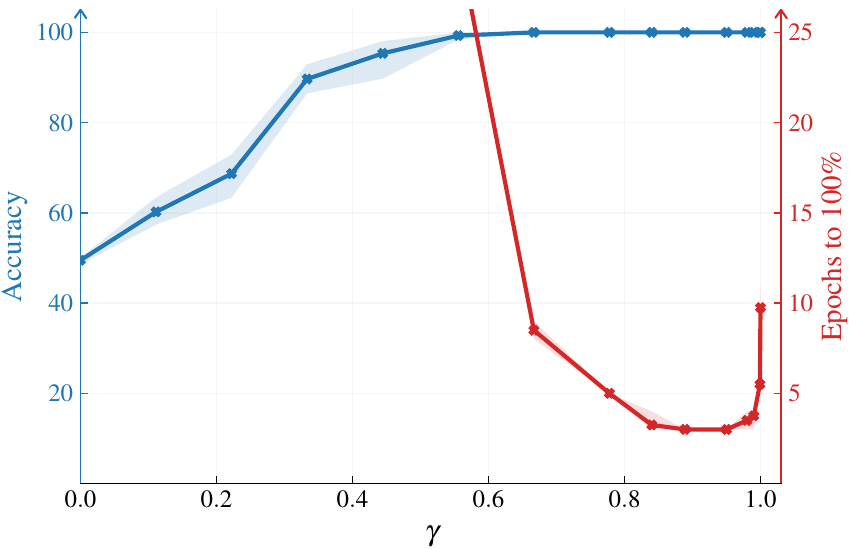}
  \caption{Impact of $\gamma$ over performance and convergence. Accuracy is shown in blue, and the number of epochs to reach a perfect accuracy is in red.}
  \label{fig:gamma_study}
\end{figure}

As our method introduces a new hyperparameter $\gamma$, we investigate its influence through an experiment on our toy dataset (see section~\ref{sec:toy_dataset}). We train a single ChaCAL layer in the exact same conditions as our other experiments on the same dataset. In particular, we only change $\gamma$ between the runs. We report our results in Figure~\ref{fig:gamma_study}.

A first observation is that, as expected, $\gamma=0$ is equivalent to the standard transformer. As $\gamma$ increases, the model gets closer to the expected behavior of ChaCAL and reaches 100\% accuracy for $\gamma \gtrapprox 0.5$.

Although ChaCAL successfully learns the task for any $0.5 < \gamma < 1$, we observe that convergence speed differs and is minimized around $\gamma=0.9$, which is the value we used throughout our experiments.

As clearly visible in Equation~\ref{eq:layer_out}, when $\gamma=1$ the output is either zero or undefined. This explains the increasing instability when $\gamma$ gets too close to 1, with the model taking more epochs to converge for $\gamma \geq 0.98$. However, we believe such instabilities could be handled by adapting other hyperparameters such as the learning rate and the $\beta_1$ and $\beta_2$ parameters of Adam.

\section{Experimental Setup}
\label{sec:ExpSetup}
\subsection{Hardware}
We ran our experiments using Nvidia A40 and A100 GPUs. Each model was trained on a single GPU -- the experiments did not require the use of distributed training schemes. We used mixed precision to speedup training.

\subsection{Training}
All models were trained using the Adam \cite{kingma2014adam} optimizer, or AdamW \cite{loshchilov2017adamw} when using weight decay. We ensured that no model overfitted during training.

All models use the standard negative log-likelihood loss for learning their tasks. Apart from weight decay, no special regularization techniques were used -- although we experimented with dropout and label smoothing.

\subsection{Hyperparameters}
We report the hyperparameters set for the different experiments in Table~\ref{tab:hyperparams}. Unless stated otherwise, we use the default values from the original papers (e.g. GPT-2 or Adam).

\begin{table*}
    \centering
    \begin{tabular}{lcccc}
    \hline
     & \textbf{Toy dataset} & \multicolumn{2}{c}{\textbf{Boxes dataset}} & \textbf{Language modeling} \\
     &  & \textbf{default} & \textbf{advanced} & \\
    \hline
    \textbf{\boldmath{$d_\text{model}$}} & 512 & \multicolumn{2}{c}{512} & 768 \\
    \textbf{\boldmath{$d_\text{inner}$}} & 2048 & \multicolumn{2}{c}{2048} & 3072 \\
    \textbf{\boldmath{$n_\text{heads}$}} & 8 & \multicolumn{2}{c}{8} & 12 \\
    \hline
    \textbf{learning rate} & 3e-4 & \multicolumn{2}{c}{3e-4} & 5e-5 \\
    \textbf{batch size} & 128 & \multicolumn{2}{c}{256} & 256 \\
    \textbf{\boldmath{$\beta_2$}} & 0.98 & \multicolumn{2}{c}{0.98} & 0.98 \\
    \textbf{weight decay} & 0.0 & \multicolumn{2}{c}{0.01} & 0.1 \\
    \textbf{training steps} & 24k & \multicolumn{2}{c}{25k} & 10k \\
    \textbf{warmup steps} & 8k & \multicolumn{2}{c}{2k} & 2k \\
    \hline
    \textbf{vocab size} & 128 & 128 & 132 & 50,257 \\
    \textbf{train samples} & $\infty$ & 1M & 500k & 550k \\
    \textbf{test samples} & $\infty$ & 10k & 5k & 1100 \\
    \textbf{runs per setting} & 4 & 3 & 4 & 1 \\
    \hline
    \end{tabular}
    \caption{Hyperparameters and dataset parameters used in our different experiments.}
    \label{tab:hyperparams}
\end{table*}

\subsection{Datasets}

\paragraph{Toy dataset} As this dataset is very easy and fast, we regenerate the entire train and test sets after each epoch, such that no example is used more than once during training. This completely discards the possibility of models overfitting.

\paragraph{Boxes dataset} This dataset is synthetic but takes time to generate. As such, we sample 500k and 1M examples for the \textit{default} and \textit{advanced} versions respectively. We use more samples for the \textit{advanced} version due to the sequences being shorter on average than for the \textit{default} version.

\paragraph{OpenWebText} As our fine-tuning experiments for language modeling are relatively fast, we only use a subset of 500k rows from the original OpenWebText dataset. As such, we perform slightly more than 2 epochs for fine-tuning each model, which is more than enough to prevent the models from overfitting.

\section{Dataset samples}

We display a random sample from each textual dataset in Tables~\ref{tab:sample_boxes_default}, \ref{tab:sample_boxes_advanced}, and \ref{tab:sample_openwebtext}.

\begin{table*}
    \centering
    \begin{tabular}{|p{15cm}|}
    \hline
    \textbf{Prompt} \\
    \hline
    The radio is in Box D, the bone and the clock and the television are in Box C, the bill and the computer and the tea are in Box B, there is nothing in Box E, the ice and the plant are in Box A, the game is in Box G, there is nothing in Box F. Move the computer and the tea from Box B to Box E. Put the milk into Box A. Put the cake into Box B. Remove the computer from Box E. Move the bone and the television from Box C to Box E. Remove the radio from Box D. Remove the bill from Box B. Remove the cake from Box B. Move the contents of Box E to Box F. Put the drug and the map into Box D. Move the contents of Box G to Box D. Move the tea from Box F to Box C. Move the tea from Box C to Box B. Move the ice from Box A to Box G. Remove the tea from Box B. Move the map from Box D to Box A. Move the drug from Box D to Box E. Move the contents of Box G to Box D. Move the contents of Box E to Box D. Remove the game and the ice from Box D. Put the cake into Box G. Put the disk into Box F. Remove the clock from Box C. Remove the drug from Box D. Remove the bone from Box F. Put the bell into Box E. Put the stone into Box F. Move the contents of Box G to Box C. Move the stone and the television from Box F to Box B. Put the magazine into Box G. Remove the disk from Box F. Remove the television from Box B. \\
    \hline
    \textbf{Expected Answer} \\
    \hline
    Box A contains the map and the milk and the plant, Box B contains the stone, Box C contains the cake, Box D is empty, Box E contains the bell, Box F is empty, Box G contains the magazine. \\
    \hline
    \end{tabular}
    \caption{Example of a prompt and the expected answer for the \textit{default} version of the boxes dataset.}
    \label{tab:sample_boxes_default}
\end{table*}

\begin{table*}
    \centering
    \begin{tabular}{|p{15cm}|}
    \hline
    \textbf{Prompt} \\
    \hline
    The television is in Box A, the cigarette is in Box B, the machine is in Box C, the cream is in Box H. Move the contents of Box B to Box F. Move the contents of Box F to Box E. Put the sheet into Box C. Remove the sheet from Box C. Put the coat into Box A. Remove the coat from Box A. Move the contents of Box H to Box F. Move the contents of Box E to Box G. Move the contents of Box G to Box E. Move the contents of Box A to Box D. Move the contents of Box D to Box G. Move the contents of Box C to Box D. Move the contents of Box G to Box C. \\
    \hline
    \textbf{Expected Answer} \\
    \hline
     Box C contains the television, Box D contains the machine, Box E contains the cigarette, Box F contains the cream. \\
    \hline
    \end{tabular}
    \caption{Example of a prompt and the expected answer for the \textit{advanced} version of the boxes dataset.}
    \label{tab:sample_boxes_advanced}
\end{table*}

\begin{table*}
    \centering
    \begin{tabular}{|p{15cm}|}
    \hline
    \textbf{Paragraph} \\
    \hline
    An earthquake of magnitude 7.4 has struck offshore near the Indonesian island of Sumatra, near Aceh province. The quake struck 214km (133 miles) south of Aceh's capital of Banda Aceh, the US Geological Survey (USGS) said. A local tsunami alert was issued and later lifted by the Pacific Tsunami Warning Center. The site is very near that of 2004's 9.2 magnitude earthquake. About 220,000 people were killed in the Indian Ocean tsunami the quake triggered. The epicentre of the latest quake was at a depth of 61.4km, about 66km (41 miles) south-west of Meulaboh district, the USGS said. The district, and other parts of Aceh, were devastated in the 26 December 2004 earthquake. Ring of Fire The quake hit at 1259 (0559 GMT). Local media reported some houses were damaged and power lines knocked down, Associated Press news agency said. The Pacific Tsunami Warning Center lifted its tsunami watch several hours after the earthquake. The earthquake caused some panic in parts of Aceh "Sea level readings indicate that a significant tsunami was not generated," the Hawaii-based centre said in a statement on its website. "Therefore, the tsunami watch issued by this center is now cancelled." The USGS earlier said it believed there was no threat of a destructive, widespread tsunami but the possibility of a local tsunami existed. Indonesia is located on the volatile Pacific Ring of Fire, a belt of tectonic activity girdling the Pacific Ocean that triggers earthquakes and volcanic activity. Aceh is on the north-western tip of Sumatra, one of Indonesia's main islands, and is frequently rocked by earthquakes. One last year near Padang in West Sumatra province killed more than 1,000 people. About 170,000 people were killed in Aceh from the 2004 earthquake and the tsunami it launched. The waves spread across the Indian Ocean to cause death and destruction as far away as Sri Lanka, Burma and Thailand.\\
    \hline
    \end{tabular}
    \caption{Sample text from the OpenWebText dataset \cite{gokaslan2019openwebtext}.}
    \label{tab:sample_openwebtext}
\end{table*}

\end{document}